\documentclass[10pt,twocolumn,letterpaper]{article}

\usepackage{cvpr}
\usepackage{times}
\usepackage{epsfig}
\usepackage{graphicx}
\usepackage{amsmath}
\usepackage{amssymb}

\usepackage[pagebackref=true,breaklinks=true,letterpaper=true,colorlinks,bookmarks=false]{hyperref}

\cvprfinalcopy 

\def\cvprPaperID{628} 

\ifcvprfinal\fi
\begin{document}

\title{Beyond Holistic Object Recognition: \\ Enriching Image Understanding with Part States}

\author{Cewu Lu$^{\star}$,~~~Hao Su$^{*}$,~~~Yongyi Lu$^{+}$,~~~Li Yi$^{*}$,~~~Chi-Keung Tang$^{+}$,~~~Leonidas Guibas$^{*}$\\
$*$ Stanford University\\
$\star$ Shanghai Jiaotong University\\
$+$ Hong Kong University of Science and Technology\\
}

\maketitle

\begin{abstract}
Important high-level vision tasks such as human-object interaction, image
captioning and robotic manipulation require rich semantic descriptions of objects at part level. Based upon previous work on part localization, in this paper, we address the problem of inferring rich semantics imparted by an
object part in still images. We propose to tokenize the semantic space as
a discrete set of part states. Our modeling of part state is spatially
localized, therefore, we formulate the part state inference problem as a
pixel-wise annotation problem.  An iterative part-state inference neural
network is specifically designed for this task, which is efficient in
time and accurate in performance.  Extensive experiments demonstrate that
the proposed method can effectively predict the semantic states of parts
and simultaneously correct localization errors, thus benefiting a few
visual understanding applications. The other contribution of this paper
is our part state dataset which contains rich part-level semantic
annotations.
\end{abstract}

\section{Introduction}
\label{sec:intro}

Recently there has been growing interest in understanding the detailed
semantics from images, because important high-level vision tasks such as
human-object interaction, robotic manipulation and image captioning
require object understanding beyond holistic object recognition. In
particular, rich description of objects at part level is necessary since
interaction among objects are often manifested as contact of the pertinent
object parts.

\begin{figure*}[tb]
\centering
\begin{tabular}{@{\hspace{0mm}}c@{\hspace{1mm}}c@{\hspace{1mm}}c@{\hspace{1mm}}c@{\hspace{1mm}}c@{\hspace{1mm}}}
\includegraphics[width=0.2\linewidth]{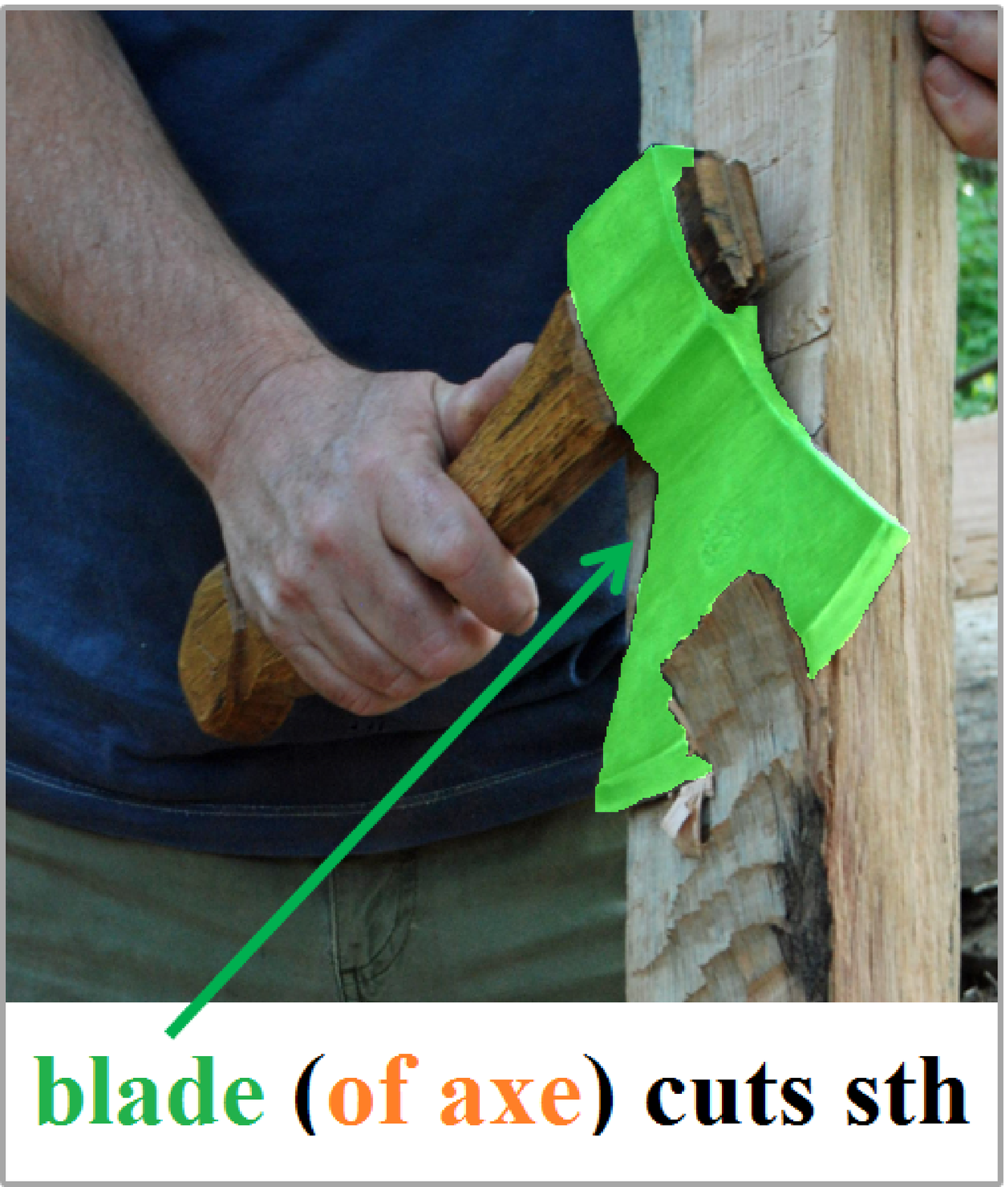} &
\includegraphics[width=0.2\linewidth]{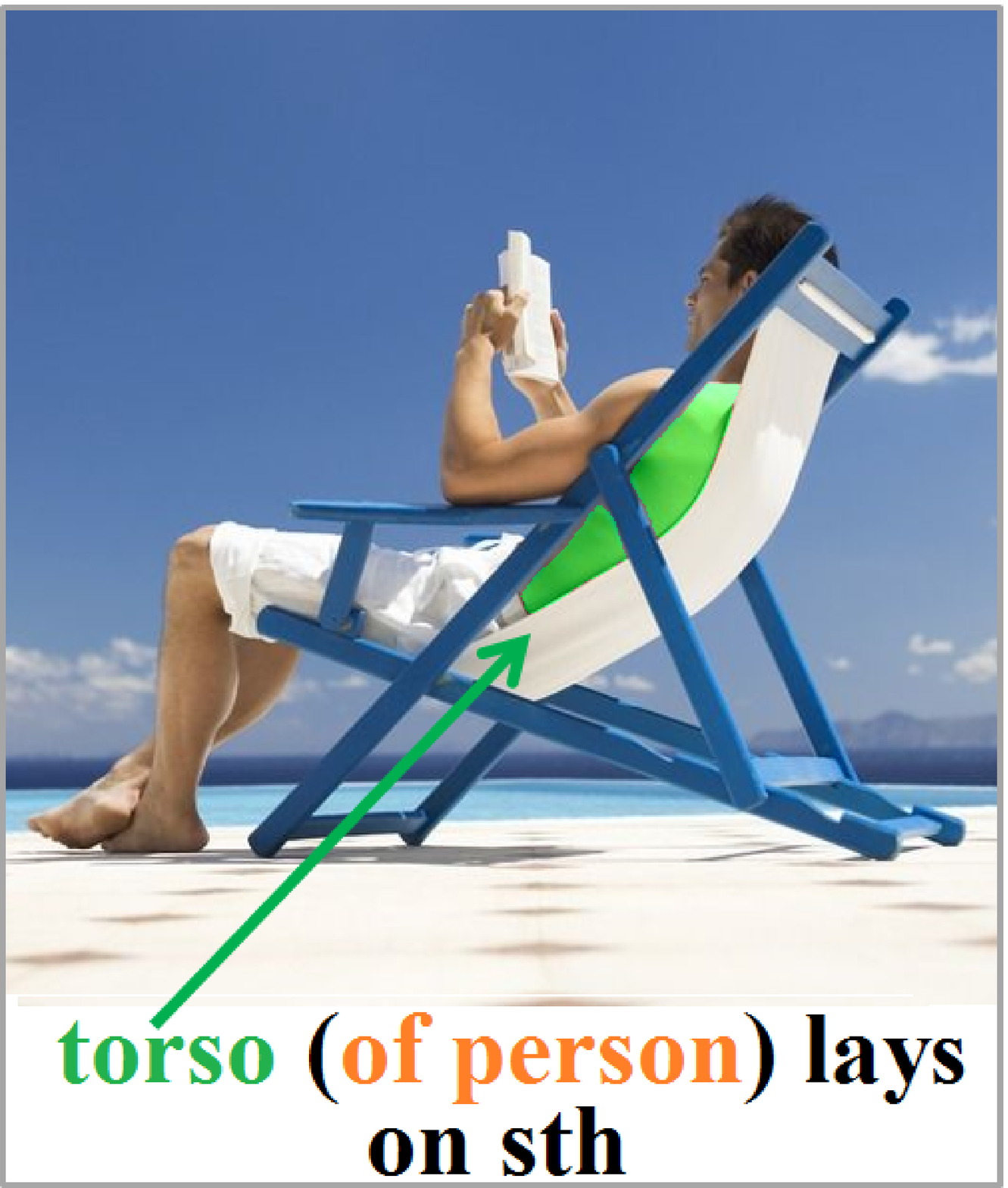}&
\includegraphics[width=0.2\linewidth]{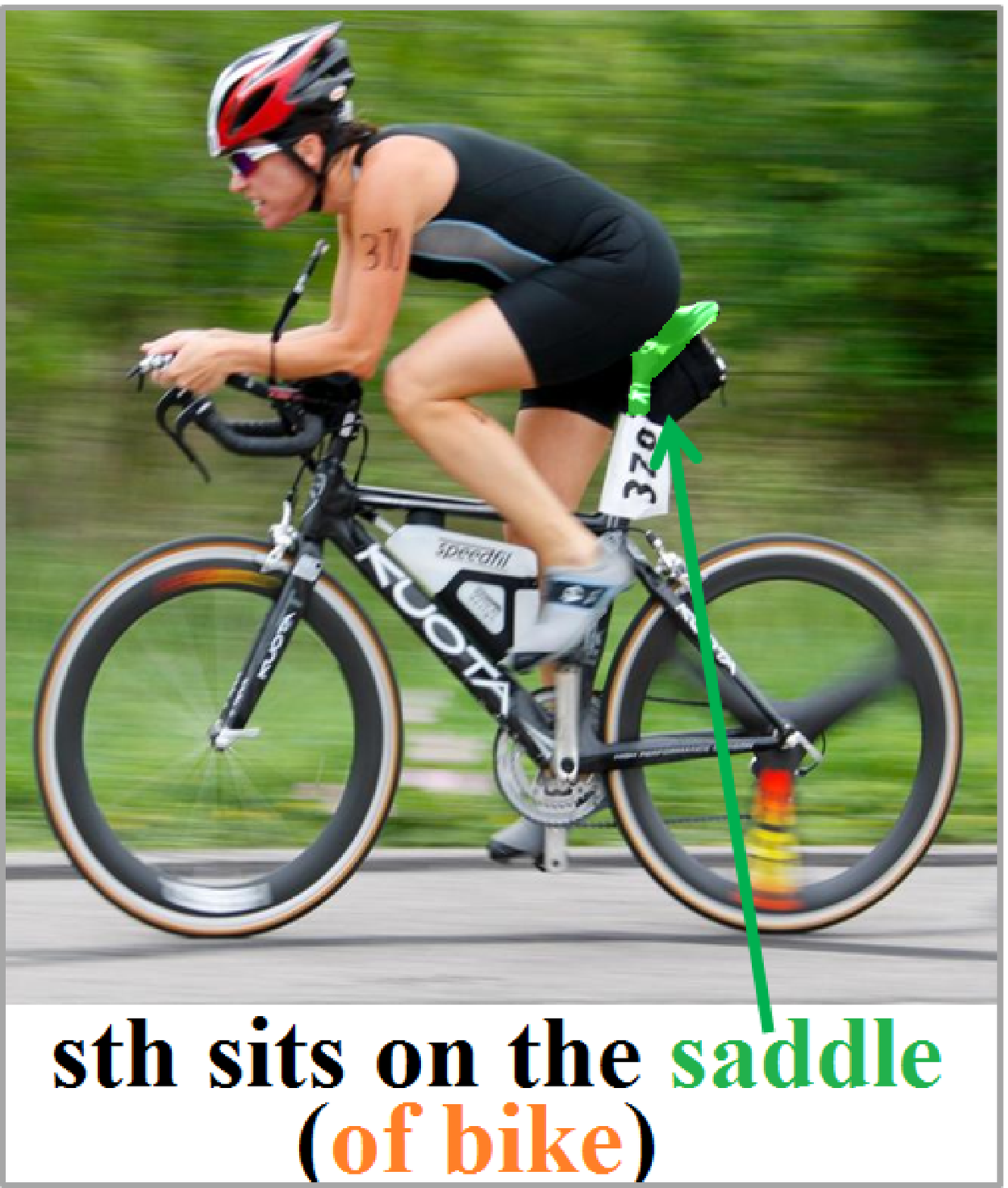}&
\includegraphics[width=0.2\linewidth]{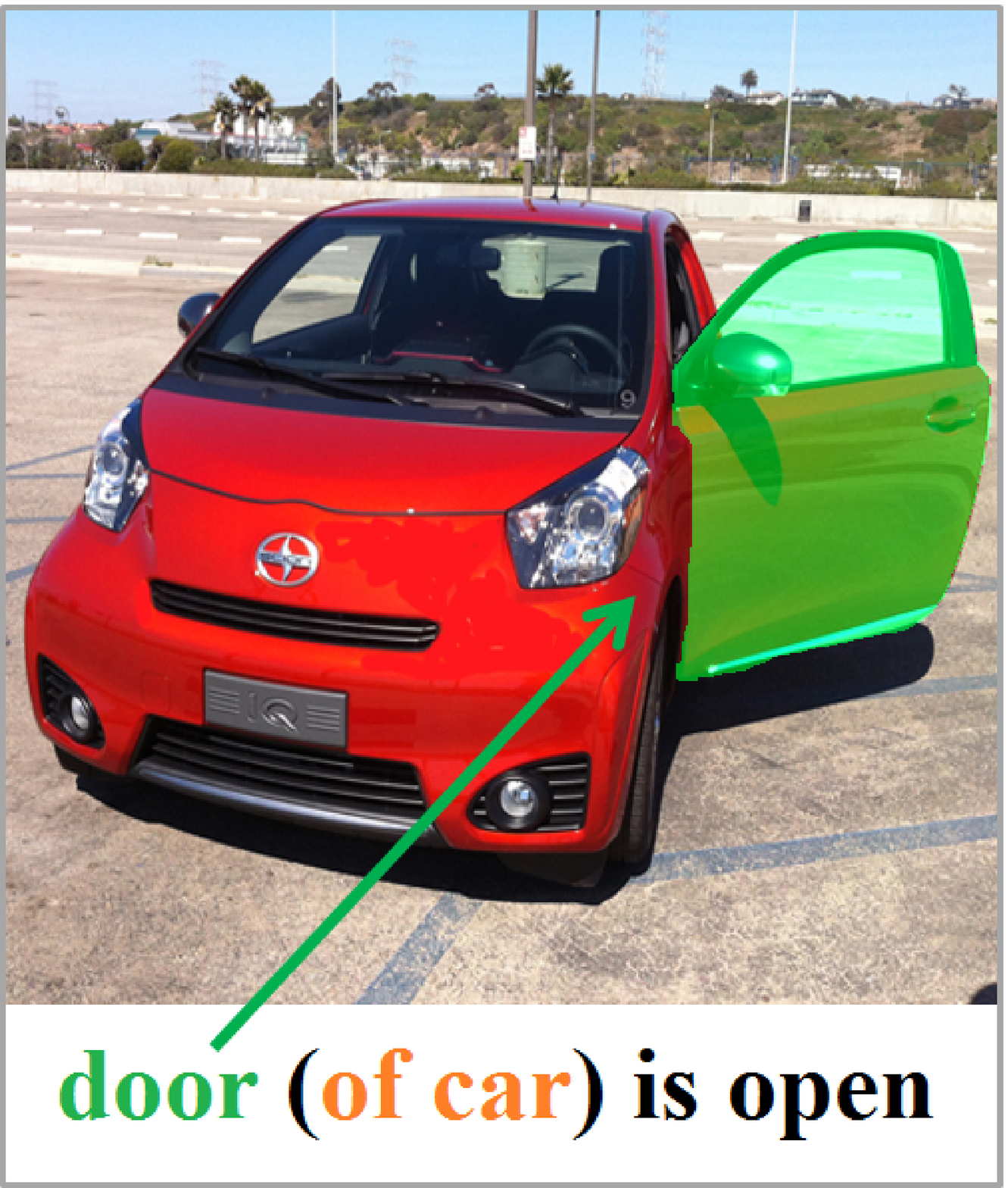}&
\includegraphics[width=0.2\linewidth]{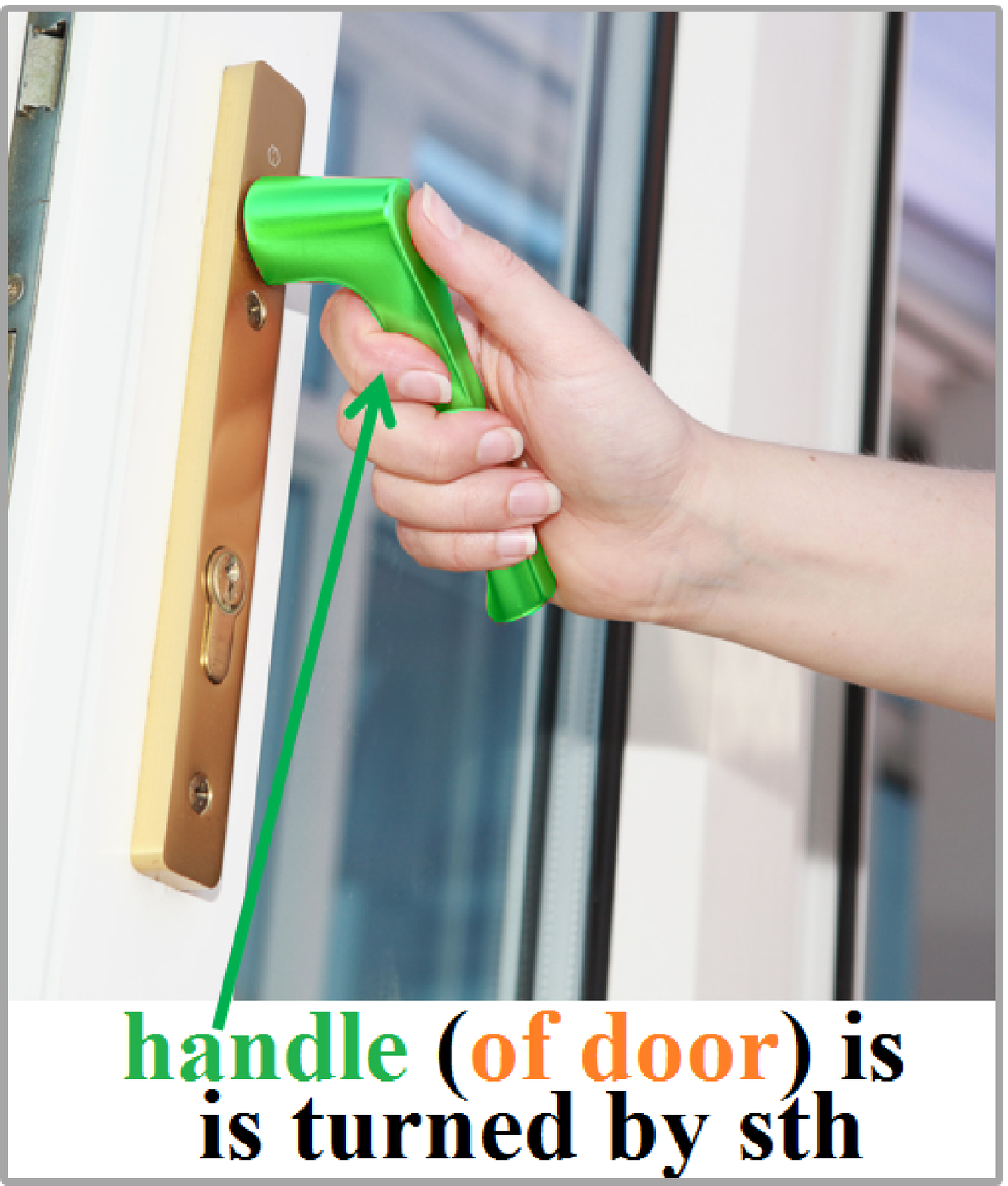}\\
(a) functionality & (b) geometry relationship& (c) affordance & (d) moment situation&  (e) interaction
\end{tabular}
\vspace{0.05in}
\caption{Parts carry rich semantics: functionality, geometry relationship, affordance, moment situation, interaction. Out-of-the-scope objects are referred to as ``sth'', short form for ``something else }
\vspace{0.05in}
\label{fig:part_state_example}
\end{figure*}

Existing work has almost exclusively focused on part localization~\cite{Wang2015Semantic}. However, localization of object parts only scratched the surface of understanding the rich information embodied in encapsulated by the object parts. In fact, through scrutinizing the visual appearance of object parts, rich semantic information about a single object and the relationships among multiple objects can be revealed.
For instance, by seeing observing a person's \emph{hand} turning the \emph{door knob}, we  infer that this person may attempts to open a door and enter a room.
Figure~\ref{fig:part_state_example} gives more examples of part semantics, ranging from including functionality, geometry relationship, affordance, moment situation (a.k.a fluent as in \cite{li2016recognizing}), to interaction. We also notice that high-level semantics on parts is important for robotic manipulation task. For example, affordance and interaction modeling are explored in \cite{shulearning} and \cite{shukla2015unified} respectively.

The paper makes a significant attempts to advance this next step: we propose
to tokenize the semantic space of relevant object parts into a discrete set of {\em part state}. Specifically,
a part of an object is associated with a set of states, each of which is characterized by
a phrase that describes its semantic meaning. Even though the discretized part states may only be an approximation to some continuous properties, in practice it is still a reasonable proxy for the representation and reasoning of objects and their interactions.

\begin{figure*}[tb]
\centering
\begin{tabular}{@{\hspace{0mm}}c@{\hspace{1mm}}c@{\hspace{1mm}}c}
\includegraphics[width=0.85\linewidth]{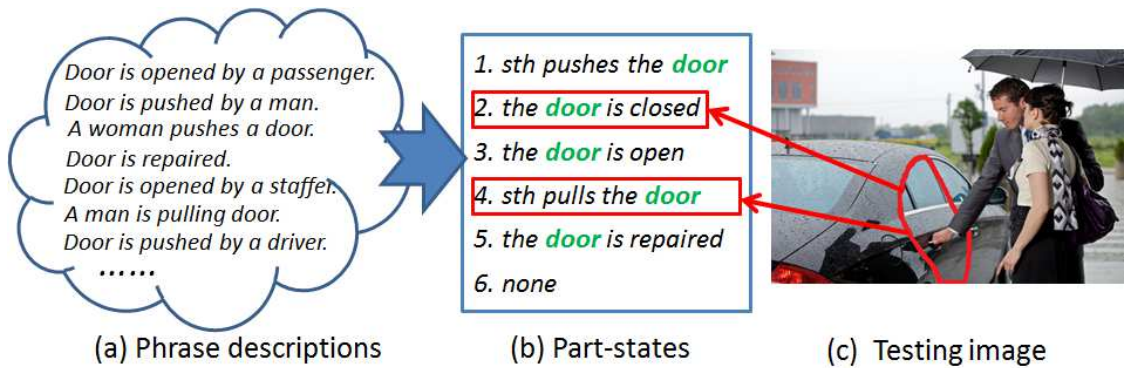} \\
\end{tabular}
\vspace{0.05in}
\caption{The pipeline of part state generation for the part door (in the car category). (a) Phrase descriptions from different annotators for different images, (b) manually summarizing of descriptions into different part states with phrases, (c) a test image falls onto two part states with indices 2 and 4.   }
\vspace{0.05in}
\label{fig:state_producing}
\end{figure*}

To implement this idea, we first have to define a vocabulary of part states, covering the common states of common objects. We address this issue by resorting to natural language processing. We collect phrase-level human descriptions on the relevant object parts in scene images, specifically, PASCAL VOC2010 images. Though simple, these phrase-level descriptions carry rich semantics (see Figure~\ref{fig:part_state_example} for examples).
The description of a particular part are readily categorized into
different discrete groups, where each group belongs to a part-state with a summarized
phrase.  Note also that our part state annotation is object centric, i.e., only the object containing the part of interest is described by its category name, all other objects are referred to as ``something else".
Figure~\ref{fig:state_producing}(a) shows
an example of part state generation, and (b)
illustrates an example of part states.

There are two desirable features associated with our part states implementation:
semantic tokenization and the ``something else" trick.  By making use of the
small semantic variation of simple parts, we can measure the semantic space on
parts with rich descriptions, and categorize them into a manageable set of
semantic tokens, or
part states. This leads to a conventional multi-class labeling problem
that can be objectively evaluated.  Therefore, our work is different from
more complex tasks such as image captioning~\cite{vinyals2015show}, where
each person has different and possibly subjective descriptions on the same
visual data.  The other feature, which is termed as the ``something
else" trick, allows attention to be paid only on target objects which
greatly simplifies the complex problem.  For non-target objects we refer
to them as ``something else''. This trick resonates with how infants learn
to grasp an object unseen before \cite{keen2011development} That is, to learn the main concept
``hand grasps something", we do not need to learn exactly what that something is.
This trick will be applied in learning our part states, which can avoid a
huge number of semantically redundant part states (e.g. ``hand grasps apple'',
``hand grasps orange'', ``hand grasps lemon", etc).

Computationally, our goal is to predict part states and simultaneously correct part
localization errors given an object image. The inherent challenge is that while
part state apparently depends on local part information, it is also
related to the holistic object appearance.  Therefore, we propose to
use an RGB-$S$ image which concatenates the input RGB image and
its part-segmented image (S). The RGB image provides holistic object
information while the part-segmented image provides local part information.
With the input RGB-$S$ image, we propose an iterative part state inference network which
iteratively optimizes the part-segmented image under the guidance
of the part state prior by minimizing the part state prediction error.
Part segment shapes and part states are closely related to each other
and thus a better part segmentation will lead to less part state prediction error.

To benchmark our performance, we construct a dataset with pixel-wise part labels and part states, which will be published alongside with the paper. Extensive experiments
show that our proposed iterative part state inference network produces excellent part
state results.

\begin{figure*}[tb]
\centering
\begin{tabular}{@{\hspace{0mm}}c@{\hspace{1mm}}c@{\hspace{1mm}}c}
\includegraphics[width=0.85\linewidth]{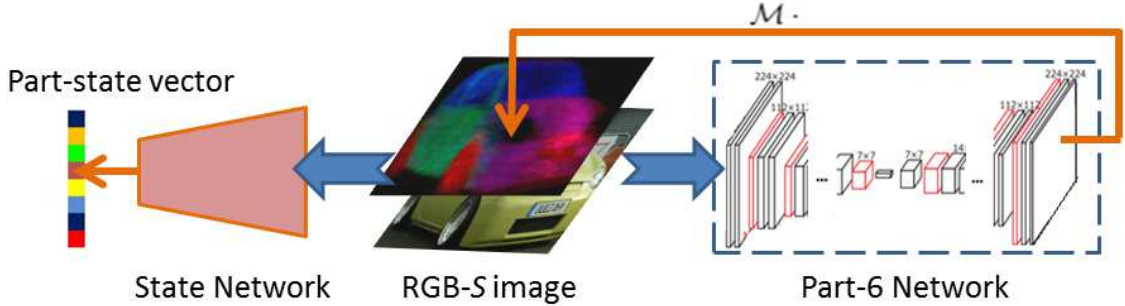}
\end{tabular}
\vspace{0.05in}
\caption{Iterative Part-state Inference  Network (ISIN) architecture. The RGB-$S$ image is the input to the part-6 network and the state network. The current $S$ image (output of the part-6 network) stacked with the input RGB image is the input RGB-$S$ image to the next iteration. }
\label{fig:ours}
\vspace{0.05in}
\end{figure*}

\section{Related Work}
\label{sec:related_work}

\paragraph{{\bf Holistic object recognition.}}
Conventional object recognition aims at object category labeling given a
test image.  Earlier work such as visual word coding~\cite{yang2009linear}
uses statistical information of local patches. The deformable part model, now
known as DPM~\cite{felzenszwalb2008discriminatively} uses part relationship
and part appearance.  Deep learning has recently made significant
contributions to object recognition. Representative network architectures
include AlexNet~\cite{krizhevsky2012imagenet}, VGG~\cite{simonyan2014very}
and ResNet~\cite{he2015deep}.  Excellent object detection methods founded on
one of these architectures include RCNN~\cite{girshick2014rich} and
Faster RCNN~\cite{ren2015faster}.  Object segmentation can be considered
as a special holistic object recognition task~\cite{long2015fully,noh2015learning}
which cuts the object from the background.

\paragraph{{\bf Layout of object parts.}} Object parts layout has been
used to provide sub-object level information.
The specific problem closely related to object parts detection is
human pose estimation~\cite{toshev2014deeppose,tompson2014joint,bourdev2009poselets}
where different human parts (e.g., head, body, hands, legs) need to be localized. In~\cite{chen_cvpr14},
a separate representation was respectively proposed for holistic object and body parts,
and a fully connected model was used to optimize their arrangement. The model
was applied to the six animal categories and achieved a better object
representation performance.  In~\cite{Wang2015Semantic}, to segment object
parts, a mixture of compositional models was used to represent the object
boundary and the boundaries of the semantic parts. This compositional model
incorporates edge, appearance, and semantic parts. The above methods localize parts only, but not in-depth explore semantics on them.

\paragraph{{\bf Image Captioning.}} Our part state can be considered as
a ``caption" on the associated part region.  Here, we survey a number of works on image captioning.  In~\cite{vinyals2015show}, a generative
model was presented that is based on a deep recurrent architecture.
Combining the recent advanced machine translation techniques, the model was
trained to maximize the likelihood of the target description sentence on the
training images.
In~\cite{Karpathy_2015_CVPR}, inter-modal correspondences were proposed
between language and visual data. To some degree image captioning explores
high-level image semantics. However, image captions vary from person to
person and are difficult to be objectively measured.

%

\section{Objerect Part-State Dataset}
\label{sec:dataset}
None of the existing datasets provides the description of part states, therefore, we build a dataset with part state descriptions for training and benchmarking our learning-based system. Our part state dataset is built on top of the part localization dataset from UCLA~\cite{Wang2015Semantic}, which provides pixelwise part membership annotation on the PASCAL VOC 2010 dataset.  We refer to our part state dataset as {\em PASCAL VOC 2010 Part State Dataset}.

Our dataset covers 15 object categories, 104,965 parts and 856 part states in total, annotated from 19,437 object images. Some parts, such as eyes and ears, are too small to detect individually, so we merge them together into one bigger part with a detectable size, e.g., eyes and ears are merged to be parts of heads. We have also fixed missing and wrong annotations. We follow~\cite{Wang2015Semantic} for the training and testing splits. We asked 15 subjects to annotate the UCLA part dataset~\cite{Wang2015Semantic} with phrase descriptions without any given constraints. Then, we manually categorize these raw descriptions into different groups where each group is indexed by a part state (with a phrase description) according to their semantic meaning. We ask different subjects to work independently. Majority rule is used to resolve different opinions when they arise. Details of part state annotation are presented in the supplementary file.

\

\section{Iterative Part-state Inference Network} \label{sec:approach}
In this section, we present the Iterative Part-state Inference  Network (ISIN) to simultaneously predict part states and part part segmentation on object images. The network operates on a novel RGB-$S$ image format. In the training phase, we learn a model from images with annotation. In the testing phase, part segments and part states on an RGB image without annotation are predicted.  We learn different models for different categories independently.  Mathematically, we use a binary variable to indicate whether a particular part state exists (i.e., 1 for exist, 0 otherwise). We concatenate all the binary variables into a ``part state vector''.  If a part is missing in an image, the vector is a zero vector. In the following, we will first introduce the part network block, and then the RGB-$S$ image and finally detail the iterative part-state inference  Network.

\begin{figure*}[tb]
\centering
\begin{tabular}{@{\hspace{0mm}}c@{\hspace{1mm}}c@{\hspace{4mm}}c@{\hspace{1mm}}c@{\hspace{4mm}}c@{\hspace{1mm}}c@{\hspace{1mm}}c}
\includegraphics[width=0.12\linewidth]{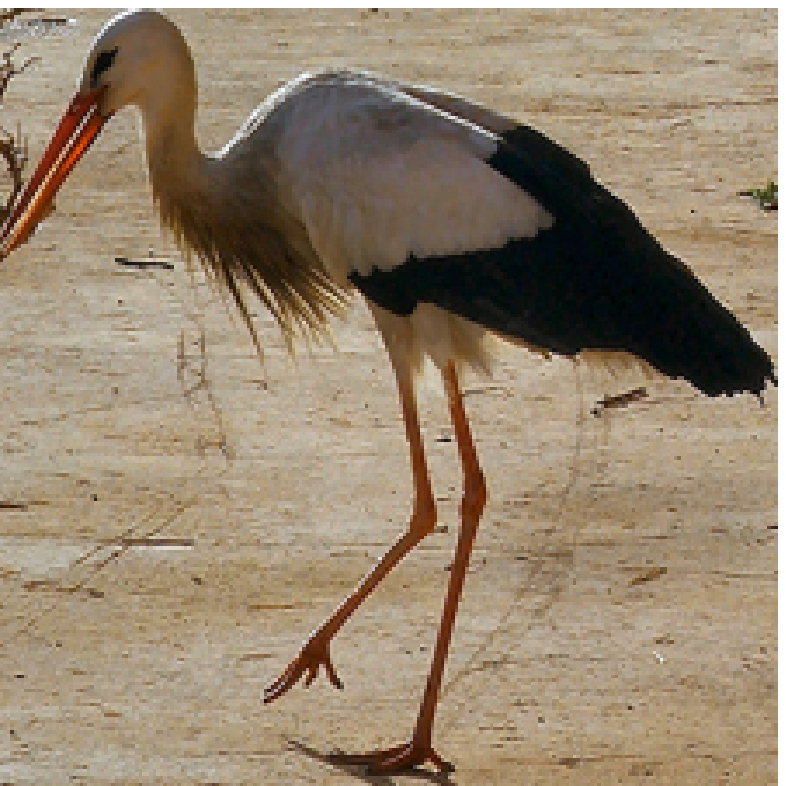} &
\includegraphics[width=0.12\linewidth]{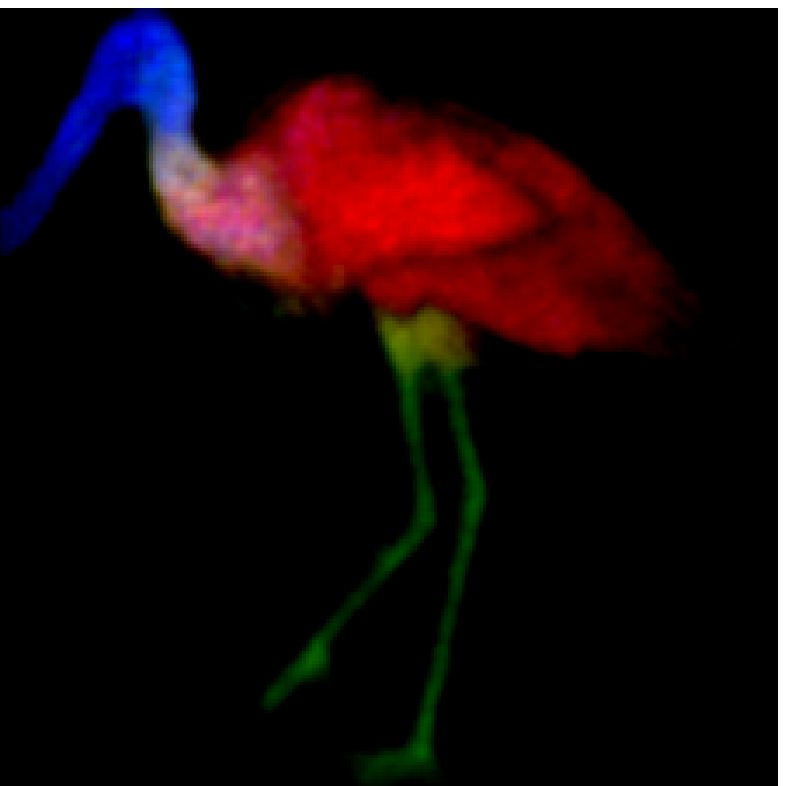}&
\includegraphics[width=0.12\linewidth]{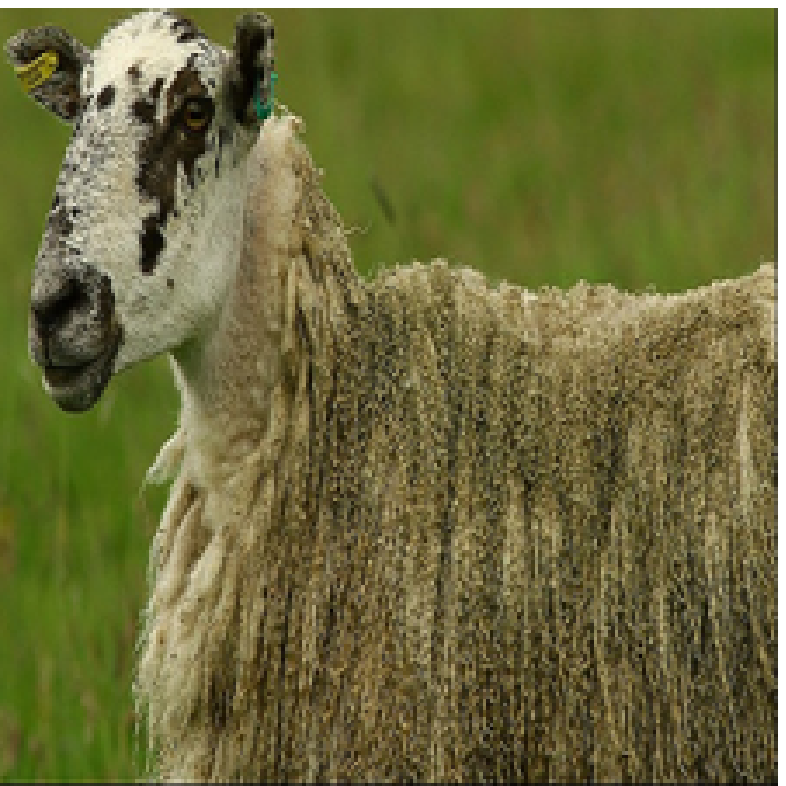}&
\includegraphics[width=0.12\linewidth]{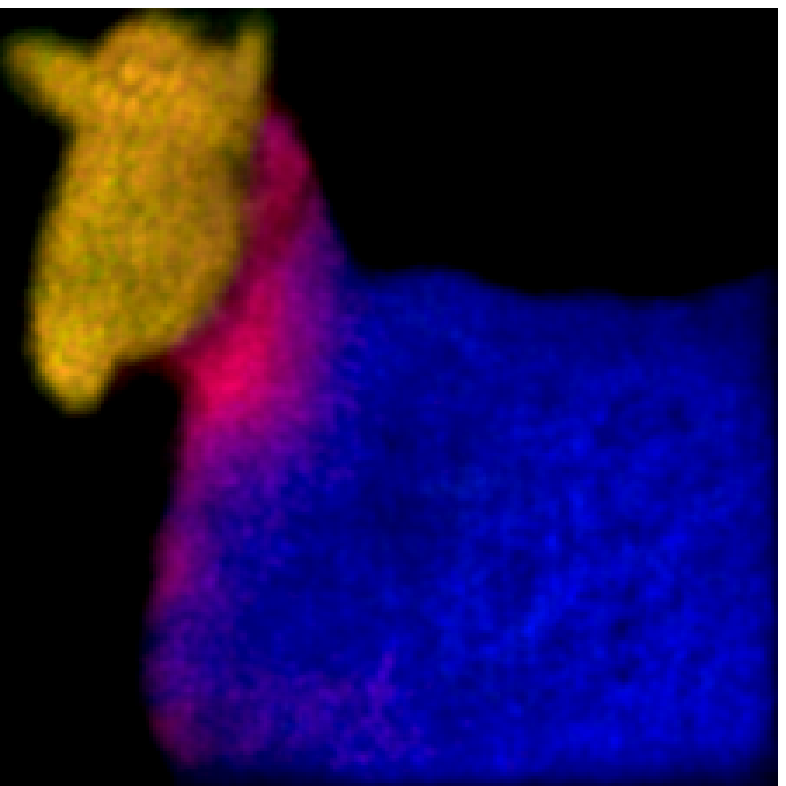} &
\includegraphics[width=0.12\linewidth]{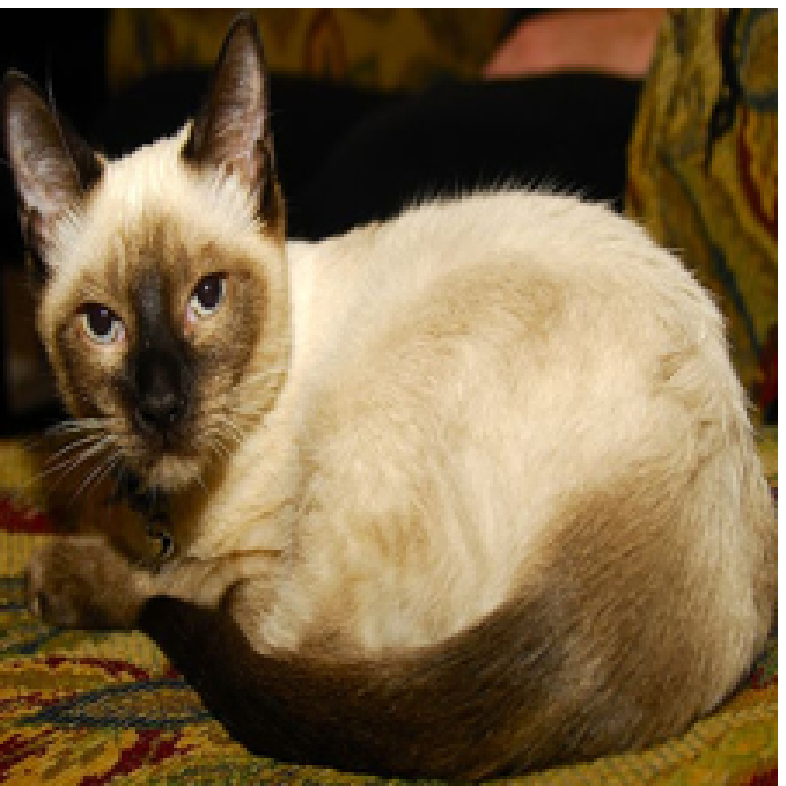}&
\includegraphics[width=0.12\linewidth]{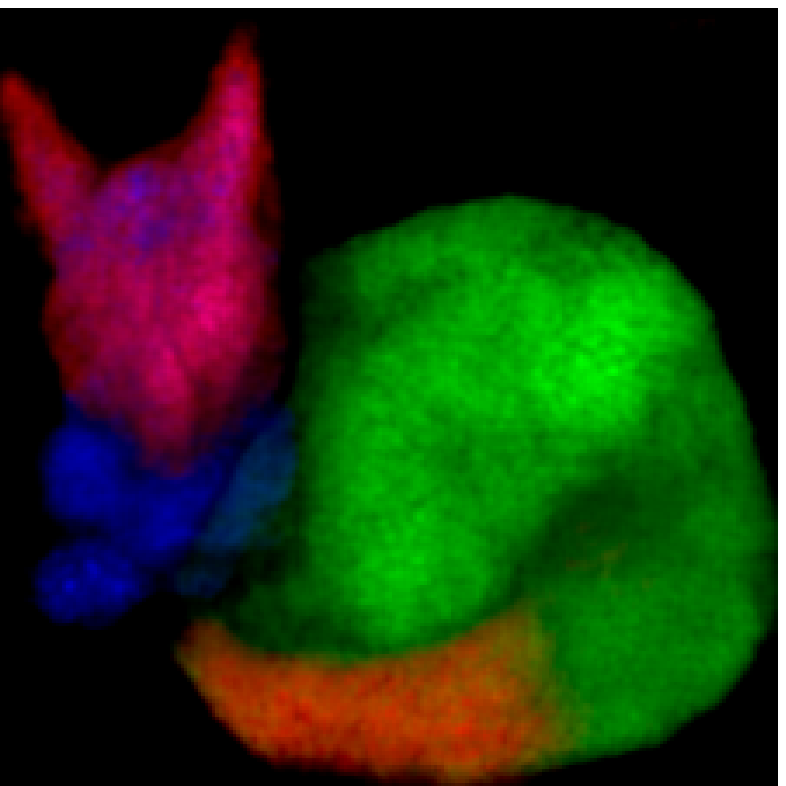}
\end{tabular}
\vspace{0.05in}
\caption{RGB-$S$ image examples. In each pair, the left is the RGB image and right is $S$ image.}
\label{fig:RGB_S}
\vspace{0.1in}
\end{figure*}

\subsection{Part Localization Network} This network aims at localizing in a pixelwise manner part regions given an object image. Therefore, the problem can be modeled as semantic segmentation. We denote the part segmentation networks with input 6 and 3 channels respectively as Part-6 network and Part-3 network. Specifically, the image input to the part network is of dimension $W \times H \times F$, where $W \times H$ are the spatial dimensions of the object image and $F$ is the number of channels, which can be 3 or 6. The segmentation solver outputs a $W \times H \times C$ volume, where $C$ is the number of part categories. The $i^{th}$ layer of the volume is a pixelwise probability map of the $i^{th}$ part class. We adopt an end-to-end deconvolution network~\cite{noh2015learning}, which is one of the state-of-the-art semantic segmentation solvers, to segment the parts.

\begin{figure}[b!]
\centering
\begin{tabular}{@{\hspace{0mm}}c@{\hspace{1mm}}c@{\hspace{1mm}}c}
\includegraphics[width=0.9\linewidth]{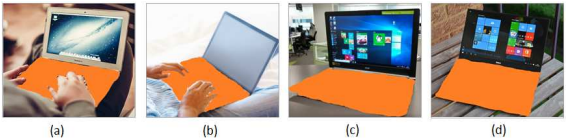} \\
\end{tabular}
\caption{The shape of a part segmentation mask is related with its part state: (a) and (b) ``keyboard is being used'', (c) and (d) ``keyboard is not being used''. In (a) and (b), keyboards are occluded by human hands, thus the occluded area is excluded from the part segmentation mask.}
\label{fig:part_state}
\vspace{0.1in}
\end{figure}

\subsection{Iterative Part Segmentation Learning}
We stack the input RGB image $I$ (resized to $224 \times 224$) and its part-segmented image $S$ (with size $224 \times 224$) to form an RGB-$S$ image (see examples in Figure~\ref{fig:RGB_S}). We denote the RGB-$S$ image as $u$ (with size $224 \times 224 \times 6$). $S$ is a 3-channel image to indicate the parts in distinct colors. Recall that the part network outputs $C$ probability maps for $C$ parts. We linearly map this volume into a 3-channel color image with a fixed mapping matrix $\mathcal{M}$ in $\mathbb{R}^{k \times 3}$, where $k$ is the number of parts. The representative colors are sampled in the color space which best discriminate among each of them. The rows of $\mathcal{M}$ are the 3D vectors of those representative colors.

At the beginning, the initial part image $S_1$ is obtained by training a Part-3 network on RGB images. With the initial RGB-$S$ image $u_1 = \{S_1, I\}$, we iteratively improve the image by implementing a Part-6 network $f(u, \Theta_f)$, which can receive a  RGB-$S$ input ($224 \times 224 \times 6$) where $\Theta_f$ is the network parameter.

In the $i^{th}$ step, the updating of $u$ can be expressed as
\begin{eqnarray}
 S_{i-1} = \mathcal{M} \cdot f(u_{i-1}, \Theta_f)\\
 u_{i} = \{ S_{i-1}; I\} \label{eq:partNetwork}
\end{eqnarray}
where $u_{i}$ is the RGB-$S$ image in the $i^{th}$ iteration, and $\mathcal{M} \cdot$ is a linear mapping operator over $f$.


The RGB-$S$ representation encodes both local parts and holistic object information: the target parts are highlighted with pixel-wise part shape to let the network look into a particular part region, while the global object appearance is revealed in the RGB image. Mapping the segmentation score volume into 3 channels reduces the computation while conveying sufficient part information.  As shown in Figure~\ref{fig:RGB_S}, we can visually distinguish different parts in the $S$ image.


\subsection{Iterative Joint Learning of Part Segmentation and Part State}
In \cite{carreira2015human}, employing feedback in building predictors is shown to be effective in handling complex structure (e.g. in pose estimation), which echoes human visual system where feedback connections are abound \cite{felleman1991distributed}. Inspired by this, we employ an iterative scheme to jointly learn part segmentation and part state. Prediction error of the part state can be considered as a feedback of part segmentation, because good part segmentation can improve part state prediction. Figure~\ref{fig:part_state} shows an example: the appearance of the keyboard provides an indication of the part state, i.e., whether the keyboard is being used or not. The part state will in turn help to guide the part segmentation (e.g., if the keyboard is not used, it should be shaped like a quadrilateral). Thus, we propose to iteratively refine part segmentation labeling under the guidance of part states which encode the knowledge of part appearance.  A better part segmentation will in turn lead to improved part states as the iterations proceed.

\paragraph{Part-state guidance} Our part state vector is predicted given an RGB-$S$ image. Denote $g(\cdot)$ as the state network whose input and output are respectively an RGB-$S$ image and part state vector. Our problem can be considered as one of multi-class labeling, so we adopt the VGG network~\cite{simonyan2014very} to solve the problem.

In the $i^{th}$ step, denote the input as $u_i$ and the output as $a_{i}$. Then, we have
\begin{eqnarray}
 a_{i} =  g(u_{i}, \Theta_g) = g(\{\mathcal{M} \cdot f(u_{i-1}, \Theta_f); I\},\Theta_g )   \label{eq:attributeNetwork}
\end{eqnarray}
where $\Theta_g$ is the network parameter.

\paragraph{Objective Function}
In the $i^{th}$ iteration, we jointly minimize the two tasks. Given $u_{i-1}$,  the objective function
$\min_{\{ \Theta_g,\Theta_f \}} G_i(\Theta_g,\Theta_f)$ is
\begin{eqnarray}
\sum_{j=0}^{N} \{  l(f(u_{i-1}, \Theta_f), \Theta_g), a_{gt}^{j}) + \lambda l( f(u_{i-1}, \Theta_f) , s_{gt}^{j})  \}
\label{eq:trainning_single}
\end{eqnarray}
where $N$ is the total number of training samples, $u_{i}^{j}$, $a_{gt}^{j}$ and $s_{gt}^{j}$ are respectively the RGB-$S$ image, ground truth part state vector and part segments of the $j^{th}$ sample; $\lambda = 0.2$ is a hyper-parameter that was obtained through grid search for maximizing the performance on the validation set.  The function $l(\cdot)$ measures the distance in the form of a soft-max loss error.  We iteratively train the model. The stopping criterion can either be the loss error being smaller than a certain threshold, or the iteration number exceeding a maximum number $M=12$. Experimental results in the following show that part state prediction is progressively refined thanks to the improvement of the part segmentation.  This is due to the fact that the network parameter of the previous iteration makes a good initialization for the network training in the current iteration. We optimize the cost function~(\ref{eq:trainning_single})  using stochastic gradient descent (SGD). Typically the iterative optimization converges in 6--8 iterations.

Given an image, in the testing phase, we iteratively compute the learned $f(\cdot)$ to produce a part segment to form the RGB-$S$ image. The number of iterations is the same as in the training phase. In the last step, part state vector is predicted based on the final RGB-$S$ image by computing $g(\cdot)$.

\begin{figure*}[tb]
\centering
\begin{tabular}{@{\hspace{0mm}}c@{\hspace{1mm}}c@{\hspace{1mm}}c}
\includegraphics[width=0.75\linewidth]{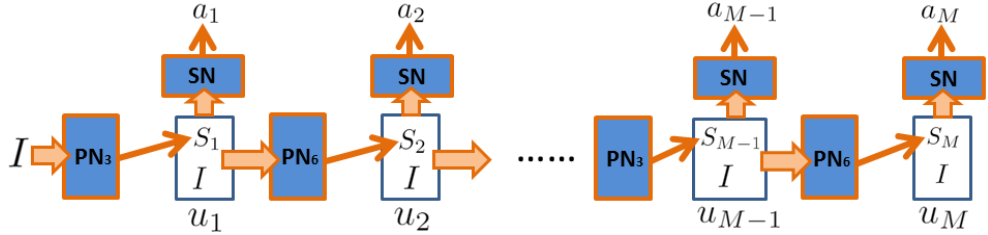} \\
\end{tabular}
\vspace{0.05in}
\caption{The unfolded network from iterations $1$ to $M$, where $PN_6$ is Part-6 network on the RGB-$S$ image and $SN$ is the state network. $PN_3$ is Part-3 network on the RGB image.}
\vspace{0.1in}
\label{fig:oursUnfold}
\end{figure*}

\paragraph{The Unfolded Architecture}
We find that our iterative framework can be unfolded into a sequential architecture as shown in Figure~\ref{fig:oursUnfold}. This unfolded architecture in fact looks similar to the recurrent neural network (RNN)~\cite{angeline1994evolutionary,medsker2001recurrent}. However, our problem is significantly different from those solved by conventional recurrent neural network and hence our resulting architecture is fundamentally different. Firstly, our data is not sequential and is not well-suited for the RNN. Second, according to the unfolded model in Figure~\ref{fig:oursUnfold}, our $S$ is analogous to the hidden units of RNN, but in RNN, the hidden units are free in the learning process, while we impose constraints on $S$ to encourage part segment formation.

Nevertheless, for the sake of comparison we also train the unfolded architecture.
Directly training the sequential objective function will lead to training a very deep model which is very time consuming. So, we train a sub-sequence iteratively. We minimize the objective function involving the error sum from $k^{th}$ to $(k+h)^{th}$ iterations. The objective function can be expressed as
\begin{eqnarray}
    \min_{\Theta_g,\Theta_f} \sum_{i = k}^{h+k} G_i(\Theta_g,\Theta_f)
\label{eq:trainning_unfold}
\end{eqnarray}
The optimization result $u_{h+k}$ will be used to train the next round which minimizes the error sum from $k+h+1$ to $k+2h$.  

Experimentally, the significant extra computation for this setup only marginally improves the performance, compared with our proposed iterative framework. One possible explanation is that the iterative architecture is already a good approximation of this unfolded architecture.

\begin{figure*}[tb]
\centering
\begin{tabular}{@{\hspace{0mm}}c@{\hspace{1mm}}c@{\hspace{1mm}}c@{\hspace{1mm}}c@{\hspace{1mm}}c@{\hspace{1mm}}c@{\hspace{1mm}}c}
\includegraphics[width=0.13\linewidth]{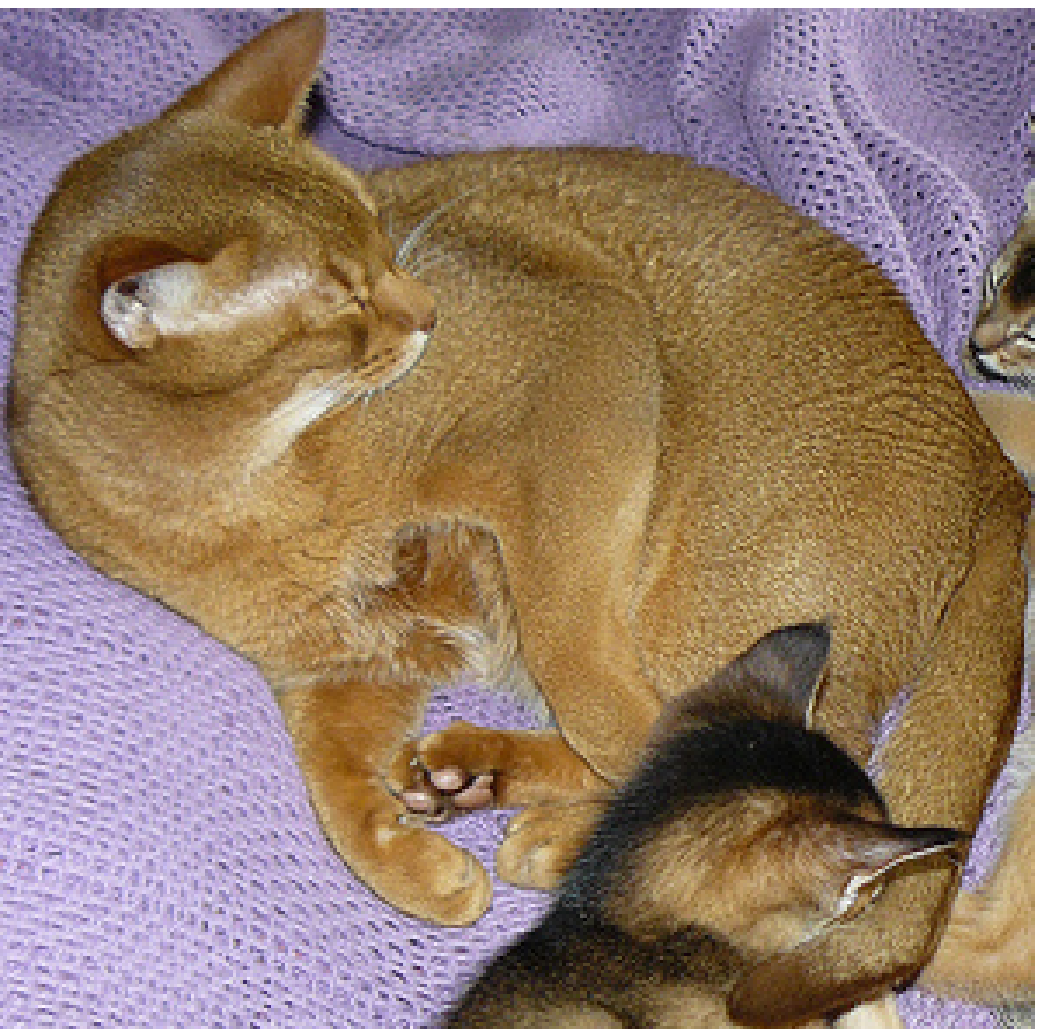} &
\includegraphics[width=0.13\linewidth]{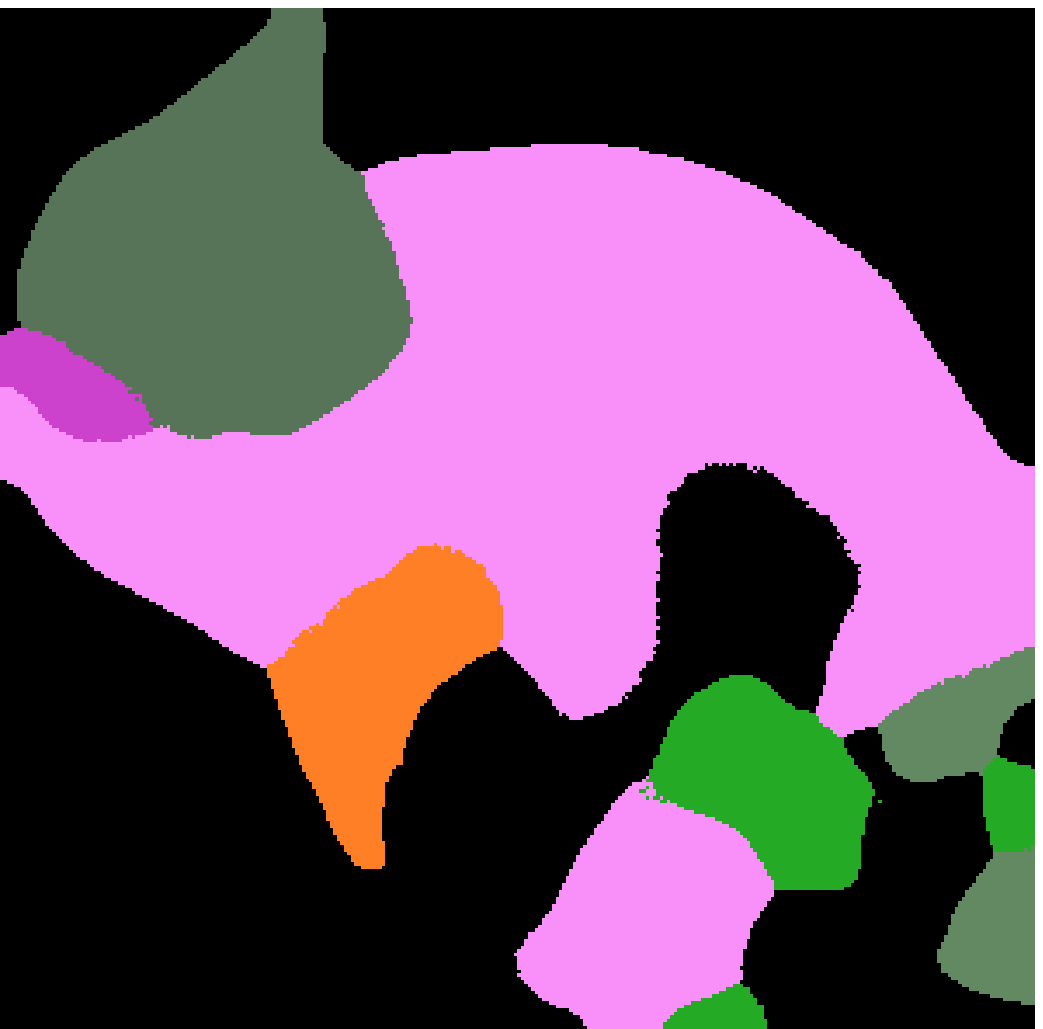} &
\includegraphics[width=0.13\linewidth]{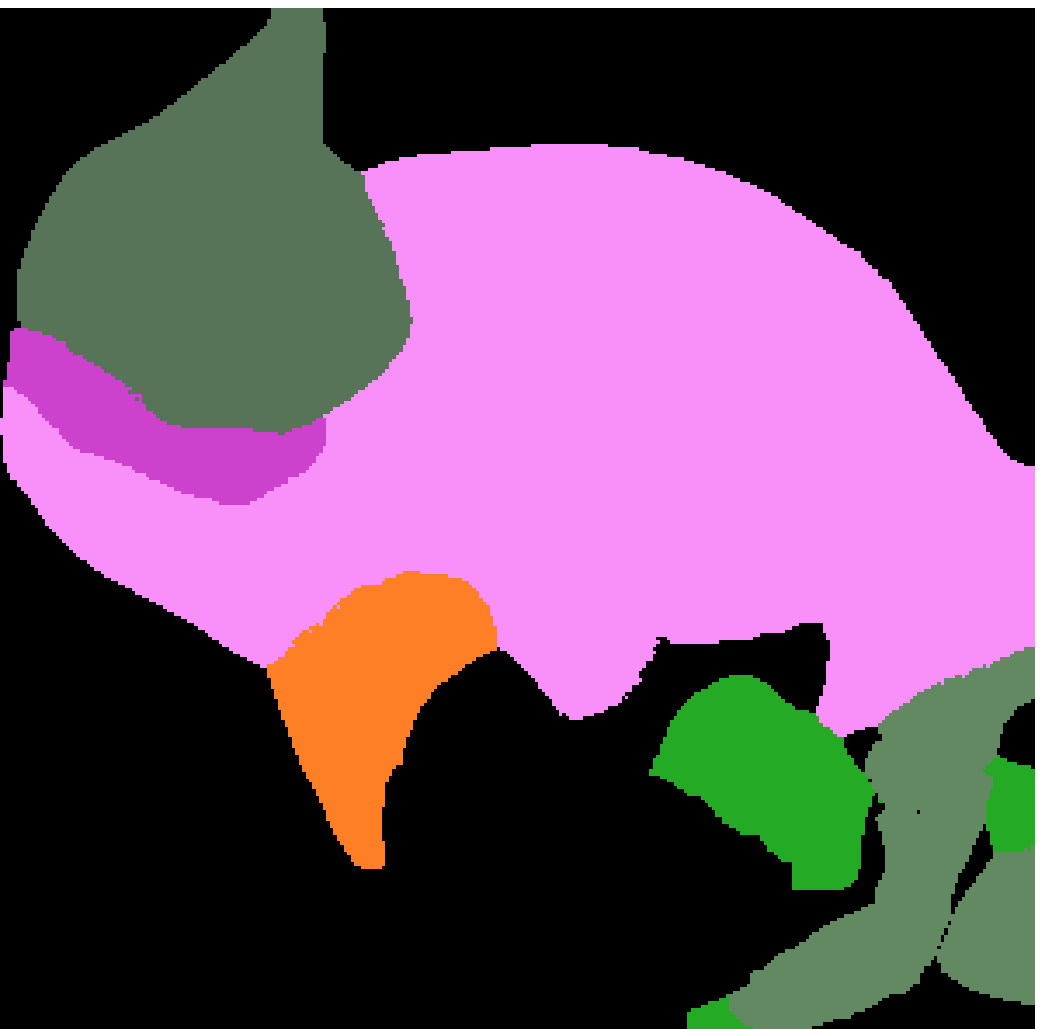}&
\includegraphics[width=0.13\linewidth]{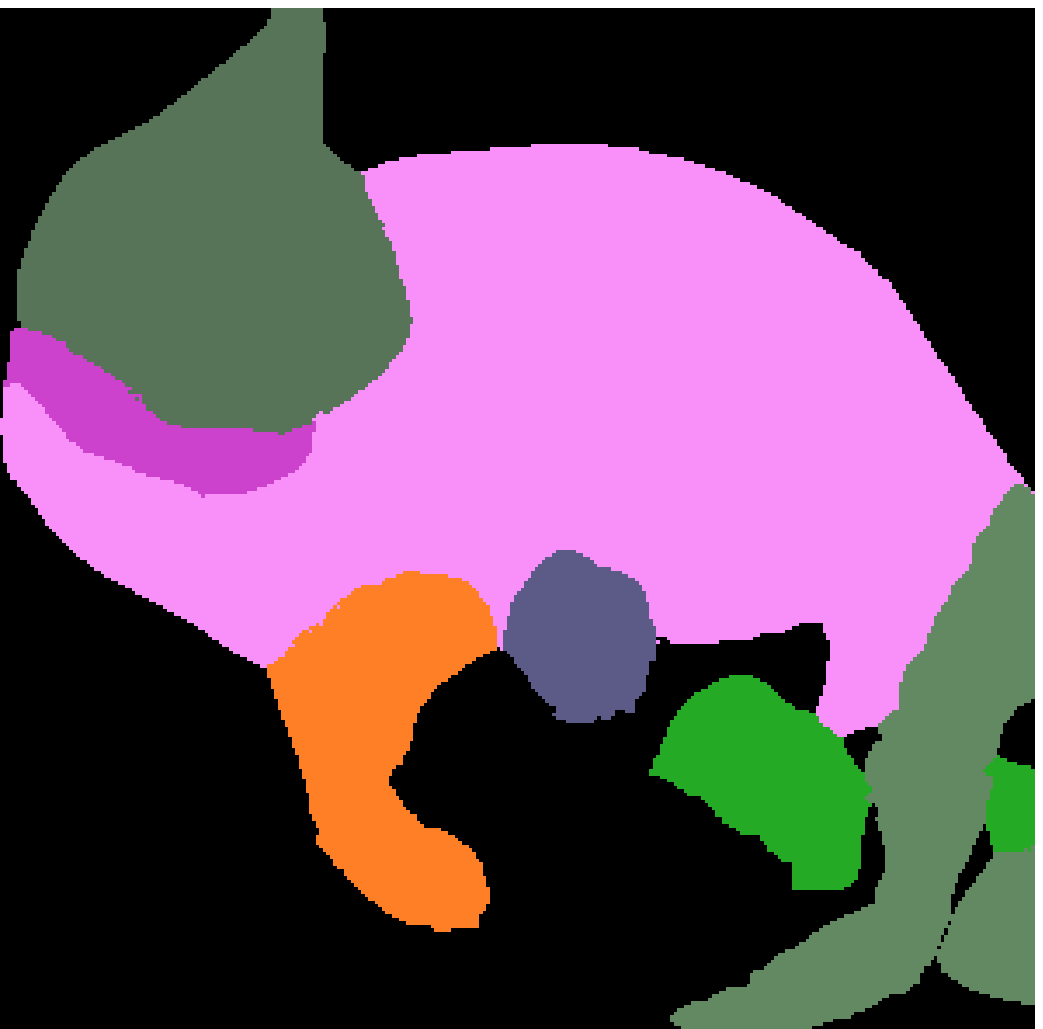}&
\includegraphics[width=0.13\linewidth]{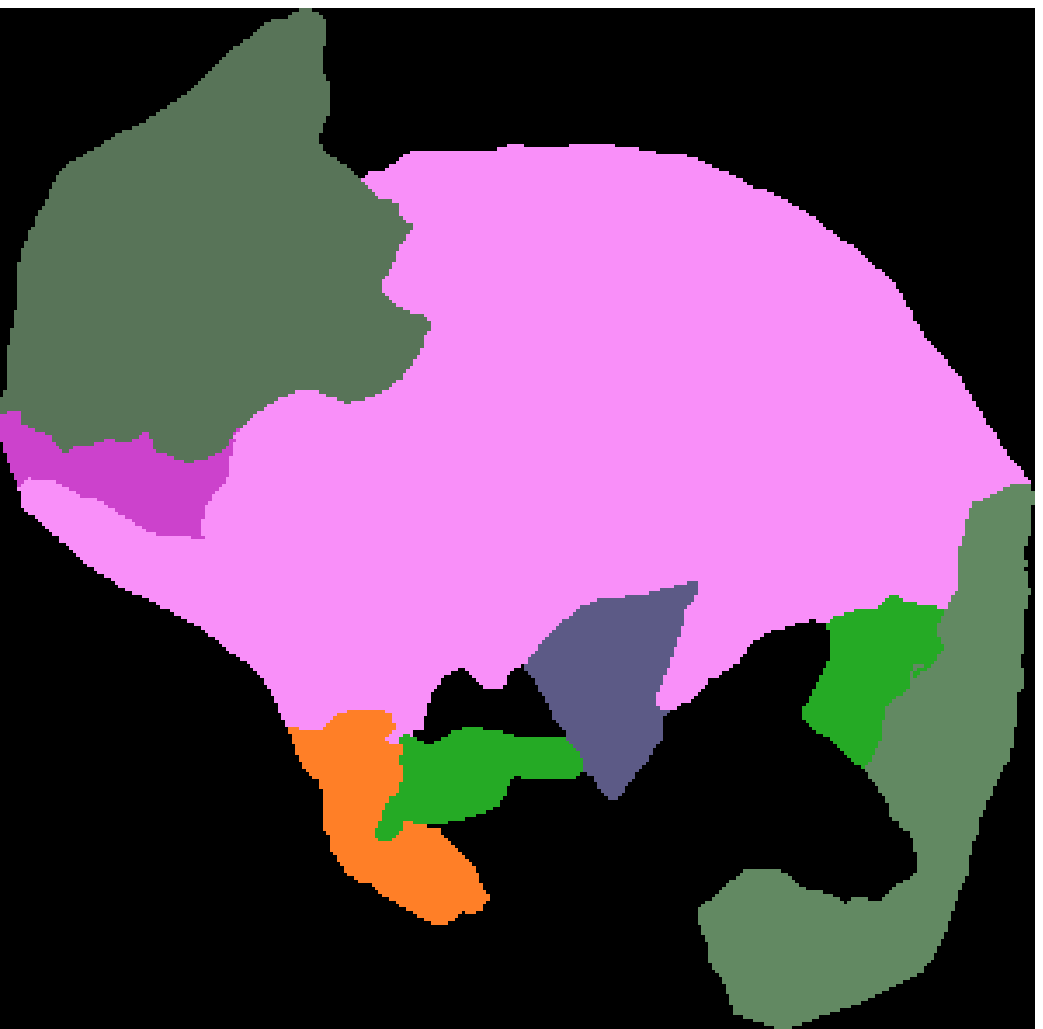}&
\includegraphics[width=0.13\linewidth]{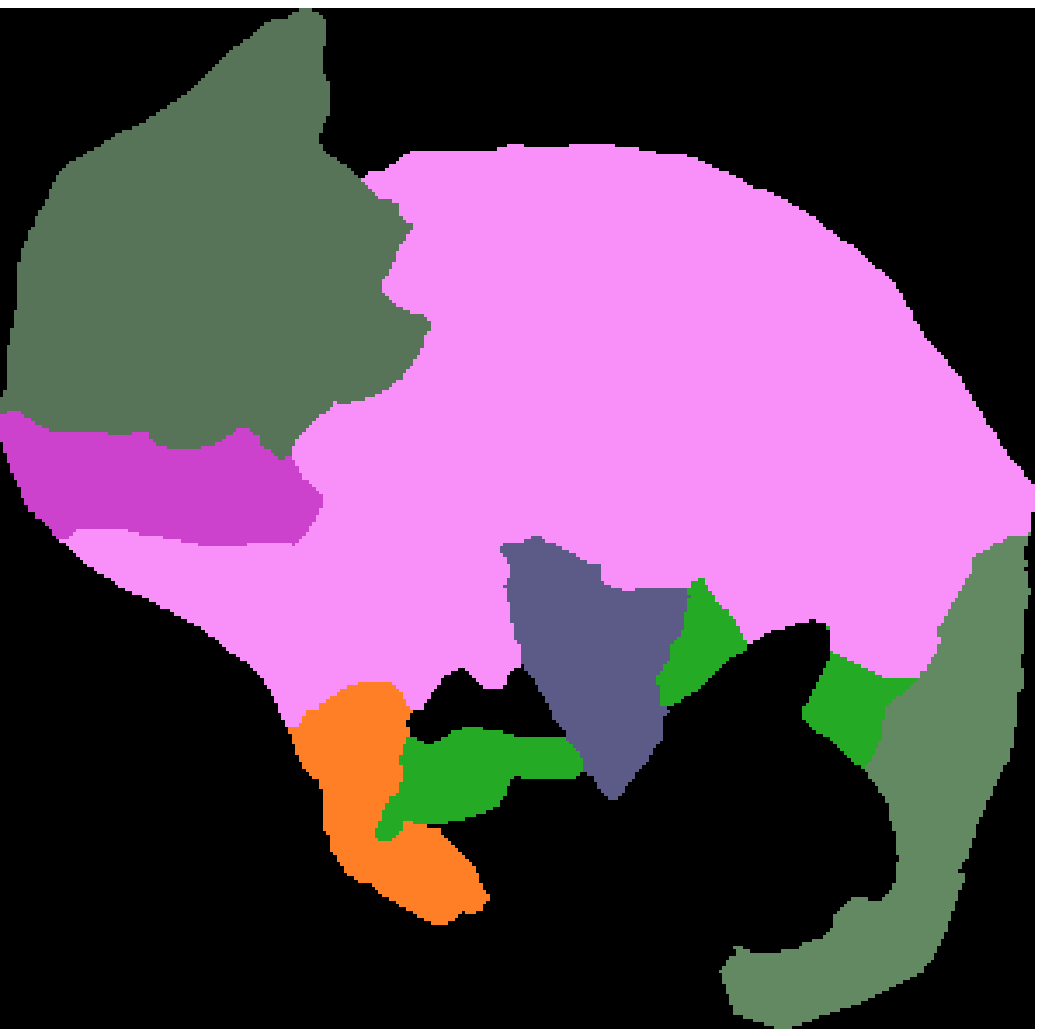} &
\includegraphics[width=0.13\linewidth]{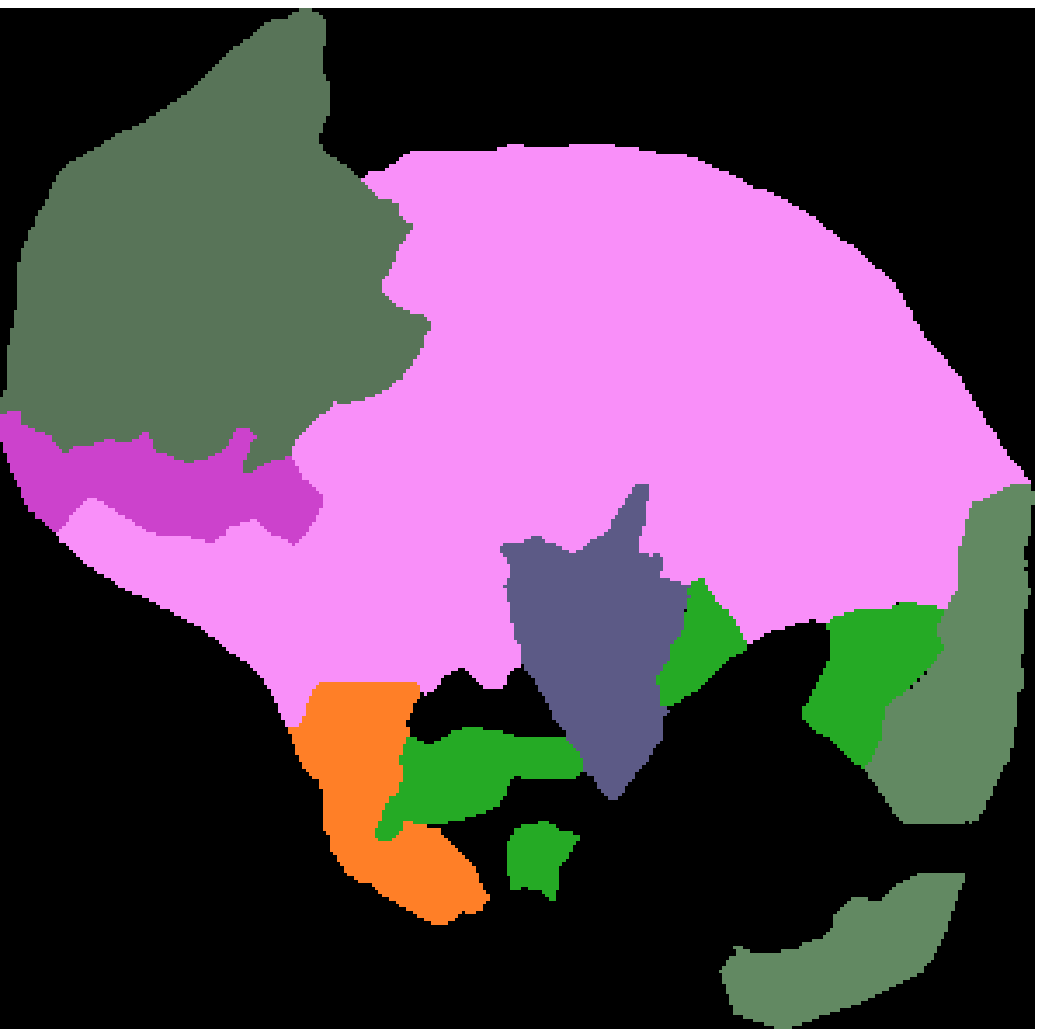}
\\
(a) input &(b) iter 1 & (c) iter 2 & (d) iter 3 & (e) iter 4 &  (f) iter 7 & (g) GT\\
\end{tabular}
\caption{The segmentation map is improved while iterations are in progress.  ``GT'' is ground truth.  }
\label{fig:seg_im}
\vspace{0.1in}
\end{figure*}

\begin{table*}
\centering
\renewcommand{\arraystretch}{1.2}
\renewcommand{\tabcolsep}{1.2mm}
\resizebox{\linewidth}{!}{
\begin{tabular}{|l|c|c|c|c|c| c|c|c|c|c|c |c|c|c|c|c |c|  }
\hline
             &aero &bike &bird &bottle &bus &car &cat &cow &dog &horse &mbike &person &plant &sheep &train & ave\\

\hline
~~~~{\bf B1} &26.7 & 10.9 & 27.0 & 24.7 & 30.2 & 28.0 & 31.8 & 24.2 & 27.3 & 24.2 & 28.8 & 28.9 & 18.0 & 26.6 & 26.4 & 25.7 \\
\hline
~~~~{\bf B2} &29.6 & 12.5 & 31.7 & 26.2 & 30.8 & 28.2 & 31.8 & 25.0 & 29.6 & 26.8 & 38.4 & 29.1 & 23.5 & 27.8 & 25.9 & 27.8 \\
\hline
~~~~{\bf B3} &32.4 & 14.6 & 34.5 & 35.4 & 38.8 & 39.8 & 35.9 & 31.7 & 32.8 & 28.8 & 36.3 & 32.3 & 25.1 & 33.4 & 32.7 & 32.3 \\
\hline
\hline
~~~~{\bf S1} &44.9 & 15.1 & 44.3 & 39.9 & {\bf 56.9 } & 49.4 & 44.8 & 38.5 & 45.8 & 37.0 & 43.0 & 45.6 & 30.3 & 42.1 & 33.1 & 40.7 \\
\hline
~~~~{\bf S2} &{\bf 52.1 } & {\bf 20.1 } & 50.3 & 39.9 & 53.5 & 58.2 & 44.3 & {\bf 45.2 } & {\bf 53.1 } & 38.0 & 53.5 & 52.0 & {\bf 37.9 } & {\bf 52.4 } & {\bf 47.0 } & 46.5 \\
\hline
~~~~{\bf S3} &45.2 & 19.8 & {\bf 50.7 } & {\bf 50.1 } & 55.2 & {\bf 59.8 } & {\bf 52.9 } & 42.2 & 42.6 & {\bf 43.2 } & {\bf 56.8 } & {\bf 55.2 } & 37.5 & 46.0 & 46.3 & {\bf 46.9 } \\
\hline
\hline
GT Seg &66.7 & 25.7 & 66.7 & 60.2 & 75.3 & 70.6 & 69.6 & 57.7 & 61.0 & 57.7 & 72.0 & 68.9 & 50.4 & 61.4 & 58.5 & 61.5 \\
\hline
\end{tabular}}
\vspace{0.05in}
\caption{The mean average precision (mAP) for different object categories on the PASCAL VOC 2010 part state dataset.
Baselines~1,2 and 3 are respectively denoted as {\bf B1}, {\bf B2}, and {\bf B3}.
Settings~1,2 and 3 are respectively denoted as {\bf S1}, {\bf S2}, and {\bf S3}.
``GT Seg'' means we use ground truth segment image as part-segmented image (S).}
\label{tab:mAP}
\vspace{0.1in}
\end{table*}

\begin{figure*}[tb]
\centering
\begin{tabular}{@{\hspace{0mm}}c@{\hspace{1mm}}c@{\hspace{1mm}}c@{\hspace{1mm}}c@{\hspace{1mm}}c@{\hspace{1mm}}c}
\includegraphics[width=0.99\linewidth]{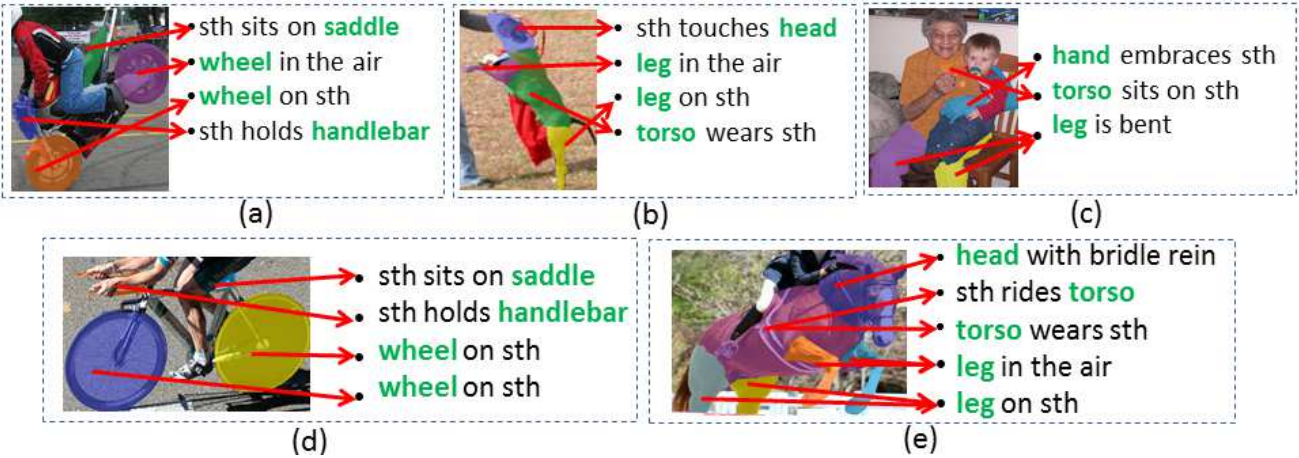}
\end{tabular}
\caption{Representative results where the detected part segments and part state on them are illustrated.}
\label{fig:qualitative}
\end{figure*}

\begin{table*}
\begin{center}
\begin{tabular}{|l|c|c|c|c|c| c|c|c|c|c|c |c|c|c|c|c |c|  }
\hline
Iteration        &  $\# 1$& $\# 2$   & $\# 3$  & $\# 4$ & $\# 5$ & $\# 6$ & $\# 7$ & $\# 8$  &  $\# 9$  &  $\# 10$  &  $\# 11$   &  $\# 12$   \\
\hline
Seg Acc          &  57.6  &  60.8    &  62.0  &   63.6 &  65.4   &   67.1 &   70.1 &    70.8 &    71.2  &   71.3    &   71.2     &   71.3    \\
\hline
mAP   &   40.7 &  41.6    &   42.5 &  43.3  &   44.0  &  44.8  &   45.6 &    45.9 &     46.2 &     46.3  &   46.3     &   46.5     \\
\hline
\end{tabular}
\end{center}
\caption{Average part state mAP (second row) and mean segmentation accuracy of part categories (first row) of as iterations are proceeding. }
\label{tab:iteration}
\vspace{0.05in}
\centering
\small
\begin{tabular}{l  c  c    c  c    c  c   }
         & \multicolumn{2}{c}{Phrase Det.}  &\multicolumn{2}{c}{Relationship Det.} & \multicolumn{2}{c }{Predicate Det.}   \\
         & R@100 &  \multicolumn{1}{c}{R@50} & R@100 &  \multicolumn{1}{c}{R@50} & R@100 &\multicolumn{1}{c }{R@50}  \\
\hline
\hline
\cite{sadeghi2011recognition}
                      & 0.05               & 0.04            &  -               & -               &  1.82          &  0.92 \\
\hline
\cite{simonyan2014very}
                      & 0.09               & 0.06            & 0.08             & 0.07            & 1.94           &  1.36  \\
\hline
B3
                      & 9.42               & 8.91            & 9.04            & 7.85            & 36.15           &  36.15  \\
\hline
\cite{lu2016visual}   & 16.32              &  15.80          &  13.01           & 12.48           &  44.19         &  44.19  \\
\hline
Ours                  & \textbf{25.37}     & \textbf{24.80}  & \textbf{26.13}   & \textbf{24.48}  & \textbf{53.50} & \textbf{53.50}  \\
\hline
\end{tabular}
\vspace{0.05in}
\caption{Results for visual relationship detection. Note that the dataset we use is different with the one used in~\cite{lu2016visual}. Here, \cite{sadeghi2011recognition}, \cite{simonyan2014very} and B3 respectively refer to baseline 1, 2 and 3.  We use the relationships with at least one object in our 15 categories.  R@100 and R@50 are respectively the abbreviations for Recall @ 100 and Recall @ 50. Note that in predicate det., we are predicting multiple predicates per image (one between every pair of objects) and hence R@100 is less than 1.}
\label{tab:relationship_results}
\vspace{0.05in}
\end{table*}

\section{Experiments}
In this section, we first introduce the evaluation metric and baseline methods for comparison,
followed by presenting a discussion.  Qualitative experiments will then be  described.
Finally, we apply our part state method on visual relationship recognition.

\subsection{Evaluation Metric} Our task consists of detecting part states with corrected localization which is analogous to object detection.  We revisit the typical evaluation metric of object detection: if the ratio intersection over union (IoU) between the predicted object box and the ground truth bounding box is larger than 0.5, and that the confidence score in the ground truth category is larger than a threshold, then we say this is a correct detection. By varying the thresholds, we can produce different precisions under different recalls. The average precisions (AP) is used to evaluate the performance.

We adopt this metric to produce a mAP measure that is reported by the mean APs over all part categories. The only difference is that the IoU we compute here is based on pixelwise segments rather than bounding boxes. We do not use bounding box, since a large number of non-compact parts simply cannot be accurately delineated using a bounding box.

\subsection{Baseline Methods}
The following baseline methods are compared:

\begin{description}
\item{{\bf Baseline 1: Global-RGB model}} We directly train a state network (using the VGG architecture~\cite{simonyan2014very}) on the RGB image to predict the part state vector. The learned model is named as global-RGB model. In the testing phase, we produce the part state vector on the input RGB image directly.

\item{{\bf Baseline 2: Local-RGB model}} We predict part states on local part regions. We use the part network (training and testing) on the RGB image to localize parts.  Then, the regions tightly bounding the parts are extracted, and we implement the VGG network on them to predict the part state vectors. The learned model is named as local-RGB model.

\item{{\bf Baseline 3: Global+Local RGB model}} We combine the previous two baseline models. The parameters of the last layer of the local-RGB and global-RGB model are respectively extracted and then concatenated to form a vector. Binary SVMs are trained on the vectors to predict the elements of the part state vector.
\end{description}

\subsection{Quantitative Evaluation}
Table~\ref{tab:mAP} compares the results of the above baseline methods and our method under different settings:

\begin{description}
\item {\em Setting 1:} Implementing one iteration only in the training. That is, the $S$ image is not iteratively updated.
\item {\em Setting 2:} Learning the model to use our iterative state-part guided network as described in section~\ref{sec:approach}.
\item {\em Setting 3:} Training the unfolded architecture model with Eq.~(\ref{eq:trainning_unfold}) (using subsequence length $h = 3$).
\end{description}

The results of baseline~1 and baseline~2 indicate that it is insufficient to model only globally (whole object image) or locally (part regions): to effectively perceive a part we must consider both the local part appearance and the object context.  For baseline~3, although local and global information are considered, they are not jointly learned, which explains the performance drop of $7\%$ in mAP when compared with our method.  In contrast, our RGB-$S$ image format can effectively derive local part regions in the pixel level ($S$ image), while providing the object appearance and its relationship among different parts (RGB image). Furthermore, our experiments verify that the iterative scheme ({\em setting 2}) outperforms the non-iterative solver ({\em setting 1}) by $5\%$. If we train on the unfolded architecture Eq.~(\ref{eq:trainning_unfold}), the improvement is very minor at $0.4\%$. A possible explanation is that iterative training ({\em setting 2}) is already a good approximation.  The drawback of the unfolded architecture ({\em setting 3}), however, is the large computation cost. In short, we recommend {\em setting 2} for solving this problem, which is a good balance between effectiveness and computation cost.

\subsection{Discussion}
\paragraph{{\bf Iterative Method}} The results in Table~\ref{tab:mAP} shows that the iterative method on RGB-$S$ image outperforms the non-iterative method.   Table~\ref{tab:iteration} tabulates the detail of iterations, where the mAP and segmentation accuracy in each iteration are shown.  An example of part segmentation during iterations is shown in Figure~\ref{fig:seg_im}. Both quantitative results and qualitative results (below) demonstrate that our iterative scheme can indeed improve the performance.

\paragraph{Influence of Segmentation}  We study the case when perfect segmentation is available
in the RGB-$S$ image, which will effectively eliminate the influence of segmentation error.
Table~\ref{tab:mAP} shows the result in the `Seg GT' row. We find that even given perfect pixelwise
part localization, we may still not be able to perfectly predict the correct part states. The
possible explanation is that the semantic meaning conveyed by the pertinent parts are
beyond simple part shape patterns.

\subsection{Qualitative Experiments}
Figure~\ref{fig:qualitative} shows representative part state prediction results.
We find that the parts are well segmented and that the part states are quite
accurately predicted, although we observe some imperfect part segmentation (see
the torso in Figure~\ref{fig:qualitative}(c)).

\subsection{Relationship Prediction}

We apply our part state method on visual relationship recognition.
We use the visual relationship dataset~\cite{lu2016visual} which includes
object-pairs relationship annotation, such as ``person holds cup''. We select
the relationships where each object-pair contains at least one object in
our 15 categories.  After manual refinement, we have a total of 6429 object-part
relationships (5000 for training, 1425 for testing) which include 31 predicate
types, such as ``hold'', ``push'' and ``under''. We found giving two object names alone is limited to infer their relationship. We should look into part level to further judge.

\begin{figure}[tb]
\centering
\begin{tabular}{@{\hspace{0mm}}c@{\hspace{1mm}}c@{\hspace{1mm}}c@{\hspace{1mm}}c@{\hspace{1mm}}c@{\hspace{1mm}}c@{\hspace{1mm}}c}
\includegraphics[width=1\linewidth]{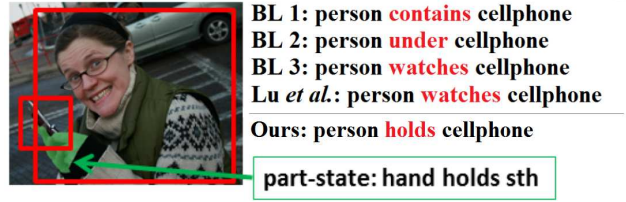} &
\end{tabular}
\vspace{0.05in}
\caption{An example on relationship prediction.  BL1 is baseline~1 (visual phrase), BL2 is baseline~2(Joint CNN), BL3 is baseline~3
(vision-language prior) and Lu \emph{et al}. refers to \cite{lu2016visual}. ``GT'' is ground truth. The part state at hand is correctly detected by the proposed method.}
\vspace{0.1in}
\label{fig:relationship}
\end{figure}

\paragraph{Solution with Part State.} The part state vectors capture rich and explicit relationship information. For example, if the part state bin for ``hand holds something'' is 1 and something is detected as a cell phone, it is straightforward to derive the relationship is ``person holds phone''. Therefore, we can use our method to extract part state vectors as feature vectors. To unify the feature length, we pad zeros for vectors with length smaller than 72, where 72
is the maximum vector length among the 15 object categories. Then, we apply SVM
to the concatenated part state vector of two objects. We combine our SVM score with the score of~\cite{lu2016visual} by simply averaging them as final relationship prediction score.

\paragraph{Baselines.} For \emph{baseline 1}, we
use the method of visual phrase recognition~\cite{sadeghi2011recognition} to classify
the relationships. For \emph{baseline 2}, we follow~\cite{simonyan2014very} to jointly
learn object and relationship in a unified CNN. For \emph{baseline 3}, we adopt the
conventional computer vision scheme: concatenate the CNN feature from VGG~\cite{simonyan2014very}
and the word-vector~\cite{mikolov2013distributed} (language prior) of two objects to
form a baseline feature.  Then, we use SVM to classify the relationship types. Approach of ~\cite{lu2016visual} is taken as the \emph{fourth baseline}.

\paragraph{Results.} The result is tabulated in Table~\ref{tab:relationship_results}, where we
follow the convention recall @ 100 and recall @ 50 in~\cite{lu2016visual}.
Recall @ {\bf x} computes the fraction of times the correct relationship
is predicted in the top {\bf x} confident relationship predictions.
The first and second baselines do not have a language prior. For baseline 3
and~\cite{lu2016visual}, the language prior improves performance. However they
preform in the holistic object level. In contrast, our method which incorporates
part level information significantly advances 12 mAP the detection performance.

\paragraph{Analysis.} Object relationship is a higher level concept than holistic objects,
and our part state encodes richer information such as interaction, affordance and functionality.
Our method shows good promise and we believe more powerful tools based on part states can be explored.
Figure~\ref{fig:relationship} demonstrates an example: given the complex concept of ``hold''
the holistic object-level appearance is not sufficient, so we should look into the key part
region -- hand.


\section{Conclusion and Future Work}
We have presented {\em part state} to tokenize the semantic space of object parts
and explore richer semantic information for image understanding. With the proposed
iterative part-state inference  Network operating on RGB-$S$ representation, we can iteratively
improve part state prediction. Extensive experiments have demonstrated the proposed
method outperforms various baseline methods. Our part state can be applied to object
relationship prediction and very  promising results are obtained.

One limitation is that the model we trained is not class-agnostic. That is,
we cannot use a unified model for all the categories. The difficulty in
training a unified, class-agnostic model stems from the fact that the output
part states for different categories vary largely among each other.  We
find about $8\%$ mAP performance drop if the model is trained without category
consideration.  Our future work is therefore to learn a unified model without
significant performance degradation in comparison to independently-learned
category-specific models.  The other limitation is that we still have no
theoretical stopping criterion on iterations. We will also perform a
principled study on the unfolded model to explain its incremental
improvement in comparison to the iterative model.

\newpage
{\small
\bibliographystyle{ieee}
\bibliography{objectpart01}
}

\end{document}